%% file: main.tex
\documentclass{article}

\usepackage[final]{corl_2020} 

\input{macros.tex}

\title{Multi-Modal Anomaly Detection for Unstructured and Uncertain Environments}

\author{
  Tianchen Ji, Sri Theja Vuppala, Girish Chowdhary, Katherine Driggs-Campbell\\
  University of Illinois at Urbana-Champaign\\
  \texttt{\{tj12,sritheja,girishc,krdc\}@illinois.edu}\\
}

\begin{document}
\maketitle

\begin{abstract}
    To achieve high-levels of autonomy, modern robots require the ability to detect and recover from anomalies and failures with minimal human supervision. Multi-modal sensor signals could provide more information for such anomaly detection tasks; however, the fusion of high-dimensional and heterogeneous sensor modalities remains a challenging problem. We propose a deep learning neural network: supervised variational autoencoder (SVAE), for failure identification in unstructured and uncertain environments. Our model leverages the representational power of VAE to extract robust features from high-dimensional inputs for supervised learning tasks. The training objective unifies the generative model and the discriminative model, thus making the learning a one-stage procedure. Our experiments on real field robot data demonstrate superior failure identification performance than baseline methods, and that our model learns interpretable representations. Videos of our results are available on our website: \url{https://sites.google.com/illinois.edu/supervised-vae}.
\end{abstract}

\keywords{Anomaly Detection, Feature Learning, Field Robots} 


\input{Sections/01-Introduction.tex} 
\input{Sections/02-RelatedWork}
\input{Sections/03-Methodology}
\input{Sections/04-Experiment}
\input{Sections/05-Conclusion}



\clearpage
\acknowledgments{This work was supported in part by the National Robotics Initiative 2.0 (NIFA\#2021-67021-33449), NSF CPS (NSF\#1739874, NIFA\#2018-67007-28379), and the Illinois Center for Digital Agriculture. We would like to thank EarthSense for providing the field robot data, and the reviewers for their thorough and constructive comments. We also thank Peter Du and Zhe Huang for their thoughtful feedback on paper drafts.}


\bibliography{BibFile}  


\input{Sections/06-Appendices}


\end{document}

%% file: macros.tex
\usepackage{amsmath,amssymb,stmaryrd,mathtools}

\usepackage{bm}
\usepackage{makecell}
\usepackage{graphicx}
\usepackage{paralist}
\usepackage{enumitem}
\usepackage{subcaption}
\usepackage{multirow}
\usepackage{booktabs}
\usepackage{wrapfig}
\usepackage{verbatim}
\usepackage[toc,page]{appendix}
\usepackage[justification=justified,singlelinecheck=false]{caption}

\usepackage[noabbrev]{cleveref}

\crefname{table}{}{}

%% file: Sections/01-Introduction.tex
\section{Introduction}
Agriculture is currently facing a labor crisis. Recent research has revealed that small, low-cost robots (agbots) deployed beneath crop canopies can coordinate to create more sustainable agro-ecosystems~\citep{Girish2019,kayacan2019tracking}. However, one big challenge for field robot development arises from the nature of highly uncertain environment in fields, where robots are likely to encounter various types of anomalies or abnormal cases. A reliable robotic system should have the ability to detect not only the presence but also the possible causes of an anomaly~\citep{christensen2008fault}. Multi-modal sensor signals can provide valuable information about robot surroundings; however, the fusion of high-dimensional and heterogeneous modalities remains a challenging problem~\citep{park2018multimodal}.

Deep-learning based anomaly detection (AD) algorithms have become increasingly popular~\citep{chalapathy2019deep}. The focus of many previous works have attempted to cast the AD problem as an one-class classification problem or as the detection of out-of-distribution samples~\citep{park2018multimodal,ruff2018deep,chalapathy2018anomaly,goldstein2016comparative}. In this work, we concentrate on multi-class classification, which potentially guides the robot to take corresponding recovery maneuvers or call for assistance. Both SVM-based and neural-network-based algorithms have been proposed for multi-class classification for the task of anomaly detection~\citep{christensen2008fault,erfani2017shared,jumutc2014multi,kim2015deep}. However, these approaches may struggle with learning multi-modal distributions~\citep{kim2015deep,Lorenz2020}, lack the representational power for high-dimensional data~\citep{christensen2008fault,jumutc2014multi}, or require a two-stage training procedure~\citep{erfani2017shared}.

Researchers often seek to reduce the dimension of high-dimensional inputs before applying the detection or classification~\citep{noda2014multimodal,rodriguez2010failure,sukhoy2012learning,kamalov2020outlier}. A simple and common approach for dimensionality compression is principal component analysis (PCA). However, several prior works have shown that reconstruction-based methods outperformed PCA in robust feature extraction~\citep{noda2014multimodal,hinton2006reducing}. A relevent approach in the reconstruction-based autoencoder domain is the variational autoencoder (VAE)~\citep{kingma2013auto,doersch2016tutorial}. VAEs learn the underlying distribution of the input data using variational inference. Recent research efforts have made noteworthy progress in leveraging the representational power of deep autoencoders in unimodal classification tasks~\citep{kingma2014semi,berkhahn2019augmenting}. Furthermore, multi-modal integration learning has been explored in out-of-distribution sample detection~\citep{park2018multimodal}, safe robot navigation~\citep{Lorenz2020}, robot behavior-learning~\citep{noda2014multimodal}, and deep reinforcement learning~\citep{chang2019robot}.

\begin{figure}[t]
  \centering
  \begin{subfigure}[b]{0.33\linewidth}
    \includegraphics[width=\linewidth]{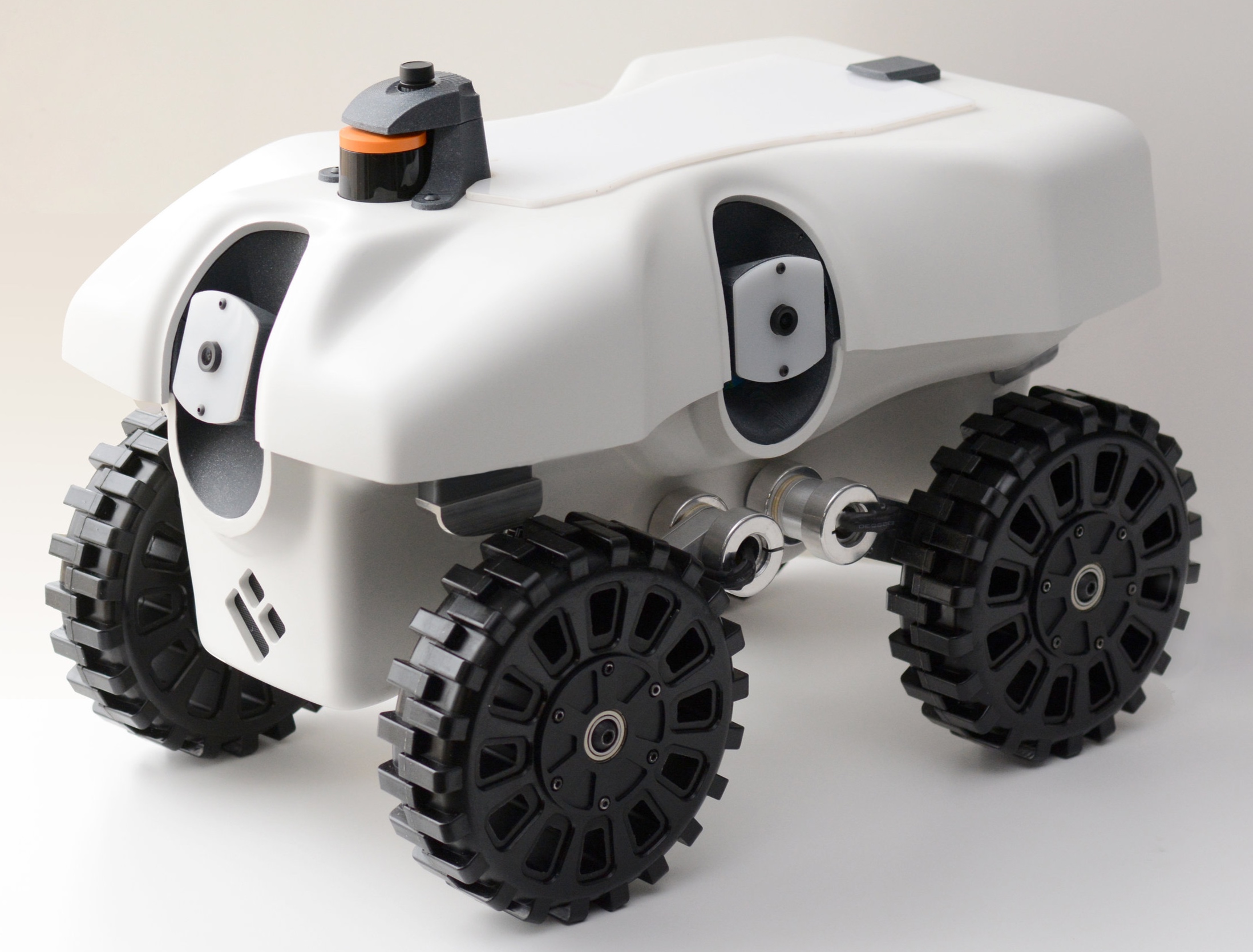}
    \captionsetup{justification=centering}
    \caption{TerraSentia robot.}
  \end{subfigure}
  \begin{subfigure}[b]{0.33\linewidth}
    \includegraphics[width=\linewidth]{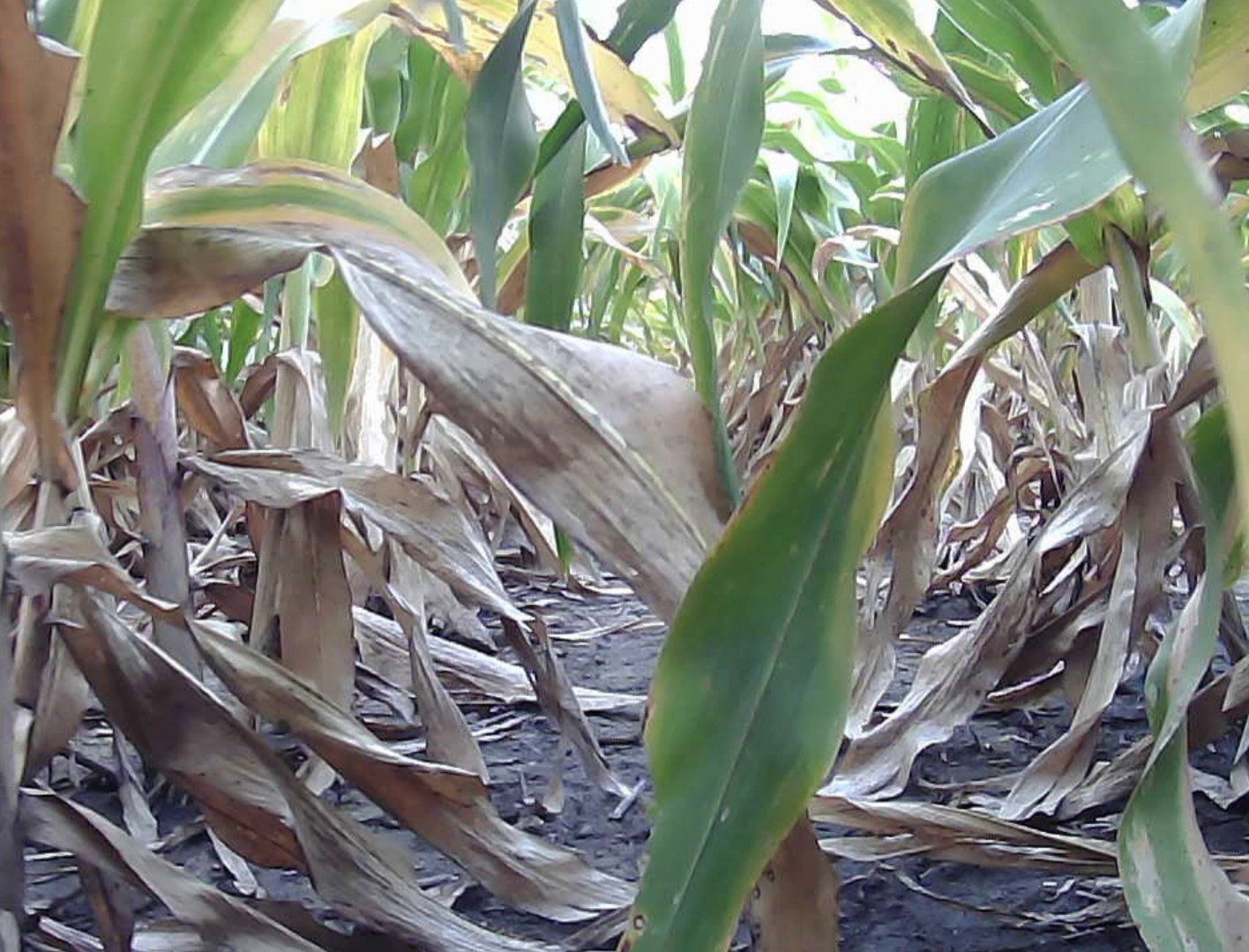}
    \captionsetup{justification=centering}
    \caption{Field environment.}
  \end{subfigure}
  \begin{subfigure}[b]{0.31\linewidth}
    \includegraphics[width=\linewidth]{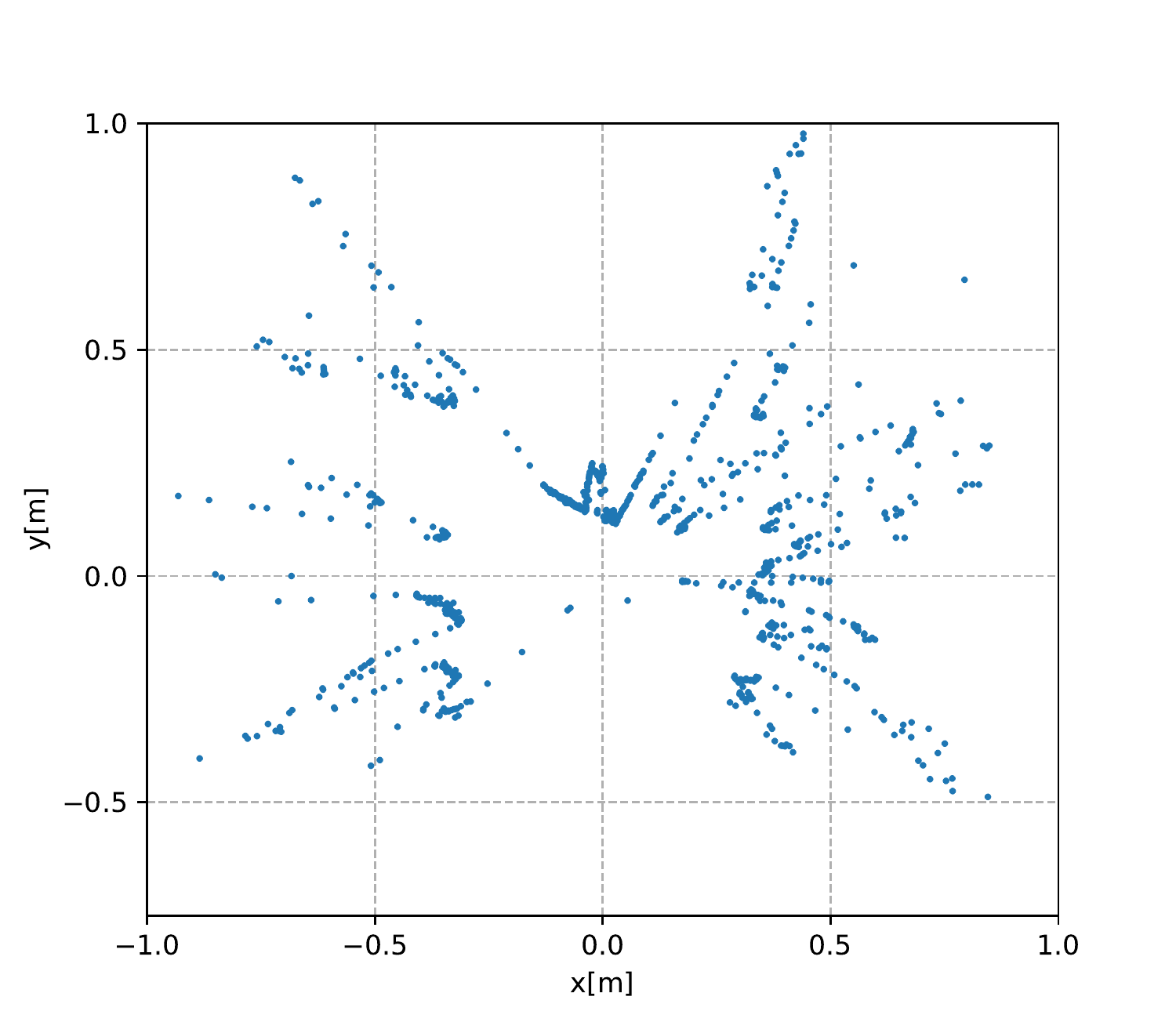}
    \captionsetup{justification=centering}
    \caption{A LiDAR scan sample.}
  \end{subfigure}
  \caption{\textbf{(a)} Our AD module is evaluated on the data collected by TerraSentia, a 15kg and 0.31m wide field robot. \textbf{(b)} A picture of robot's surroundings from the front camera in a corn field. Low-hanging leaves or weeds may block the LiDAR and camera view as the robot navigates under canopy. \textbf{(c)} A LiDAR scan sample in the dataset. The origin represents the robot position.}
  \label{fig:intro}
\end{figure}

In this paper, we approach the anomaly or failure detection problem using a multi-class classifier, which simultaneously monitors the emergence of robot failures and finds possible reasons behind failures. We use a compact agricultural robot, TerraSentia, as our testbed for our model (Figure~\ref{fig:intro}). Low localization accuracies between crop rows and uncertainties in sensory signals (e.g., LiDAR) due to weeds and hanging leaves in field environments lead to a higher failure rate than in indoor or static environments.

We propose the use of a supervised variational autoencoder (SVAE) model for the failure identification task, which utilizes the representational power of VAE to perform multi-class classification for multi-modal input modalities. Through unifying the optimization of the generative model and discriminative model in a single objective function, our method yields a one-stage training procedure. We demonstrate higher classification accuracy than baselines and interpretable learning results on data collected by the field robot.

%% file: Sections/02-RelatedWork.tex
\section{Related Work}
\label{sec:relatedwork}
Anomaly detection, also called novelty detection or outlier detection, is an important problem that has been researched within diverse application domains~\citep{chandola2009anomaly}. In robotics, the AD problem is also related to failure detection or fault detection and an anomaly detector is often defined as a method to identify when the current execution differs from past successful experiences~\citep{park2018multimodal}. Early research efforts often make use of robot or sensor models for the AD task. Multiple Model Adaptive Estimation (MMAE) uses a bank of Kalman filters to predict the outcome of several faulty patterns. A neural network is then trained to identify the current state of the robot based on the residual between the predicted readings and the actual sensor readings~\citep{goel2000fault}. 
Soika examines inconsistencies between different sensors' statements using probabilistic sensor models to detect failures~\citep{soika1997sensor}. Vemuri et al. models the faults as nonlinear functions of the measured variables in system dynamics where anomalies are declared whenever off-nominal system behaviors are observed~\citep{vemuri1997neural}. These approaches were designed for anomalies that may affect the system or sensor models; however, the generalization of these model-based methods to anomalies caused by external factors (e.g., environmental interference) remains unclear.

In the domain of supervised deep anomaly detection, deep architecture with support vector data description (SVDD)~\citep{kim2015deep} uses hidden layers with SVDD to capture features that are helpful for classification. Nevertheless, Deep SVDD shows inferior performance when compared to deep autoencoders in terms of learning the joint distribution over input data due to its unimodal assumption~\citep{Lorenz2020}. Recently, Erfani et al. proposed to perform subspace projection and landmark selection to extract robust features from the data before training the classifier. The proposed method is capable of learning complex probability distributions of high-dimensional data. However, the optimization of the feature extractor and classifier are separate and thus training is a two-stage procedure~\citep{erfani2017shared}.

Thanks to the representational power of deep autoencoders, VAEs have been widely used in AD tasks. LSTM-VAE replaces the feed-forward network in VAEs with LSTMs to learn joint distributions of multi-modal observations and their temporal dependencies~\citep{park2018multimodal}. The reconstruction probability, which considers the variability of the variable distributions, is used as a new anomaly score to detect anomalies~\citep{an2015variational}. Conditional VAEs have also been applied to identify anomalous samples given associated observable information in specific applications~\citep{pol2019anomaly}.

VAE-based methods have also been proposed to perform multi-class classification tasks. Kingma et al. propose M1 and M2 models for semi-supervised learning problem~\citep{kingma2014semi}. However, M1 requires two-stage training, and M2 assumes two independent encoder networks for labels and other latent variables, which may not lead to an optimal solution for supervised learning tasks. Semi-supervised VAEs, which can also be used for supervised learning, connect a classification layer to the topmost encoder layer providing unified training~\citep{berkhahn2019augmenting}. Nevertheless, without the use of a deep architecture for the classification layer, it is unclear how well the model generalizes to multi-modal and high-dimensional input modalities. In this work, we utilize the representational power of VAEs to improve the classification accuracy for multi-modal inputs with a one-stage training procedure.

%% file: Sections/03-Methodology.tex
\section{Methodology}
\label{sec:methodology}
Our goal is to detect and classify failures or anomalies that occur during robot navigation in the field. We introduce a general data-driven anomaly detector architecture, which can be deployed on the robot and be trained efficiently based on variational inference.

We construct our anomaly detector as a deep supervised multi-class classifier. We assume that the sensor data is multi-modal, consisting of a high-dimensional input modality $\mathbf{x}_h \in \mathbb{R}^H$ (e.g., LiDAR) and a low-dimensional input modality $\mathbf{x}_l \in \mathbb{R}^L$ (e.g., wheel encoders). The AD module outputs zero when the robot is in normal operation, and outputs non-zero values when an anomaly occurs. We further divide the anomalies to $C$ sub-classes to facilitate the robot to take corresponding recovery maneuvers for different types of anomalies or failures. At each time step, the AD module maps from a set of current sensor data $\mathbf{x}_t \in \mathbb{R}^{H+L}$ to a corresponding class label $y_t \in \{0,1,\dots,C\}$.

Our anomaly detector is composed of two parts: a feature generator (FG) and a classifier. A schematic overview of the AD module is depicted in Figure~\ref{fig:architecture} (left). We define the FG as a function $g: \mathbf{x}_h \mapsto \mathbf{z}$, which maps the high-dimensional inputs $\mathbf{x}_h \in \mathbb{R}^H$ to some latent variables $\mathbf{z} \in \mathbb{R}^d$. In most cases, we prefer $d \ll H$ to ease the training of the classifier. With proper selection and training of the FG, the model is expected to learn meaningful and robust representations / features of the original inputs in the latent space.\footnote{The FG can be implemented with fully connected layers of a neural network, or with a fully-convolutional neural network when dealing with image-like inputs~\citep{Lorenz2020}.}

The latent variables $\mathbf{z}$ generated by the FG, along with other low-dimensional inputs $\mathbf{x}_l$ are then used as inputs to the classifier: in our case, a feed-forward network. Here, we do not constrain the low-dimensional inputs $\mathbf{x}_l$ to be the raw sensor data, instead $\mathbf{x}_l$ can also involve modalities derived from the raw data which may help the classifier make the correct inference~\citep{Lorenz2020}.

\subsection{Supervised VAE Architecture}
\begin{figure}[t]
  \includegraphics[width=0.95\linewidth]{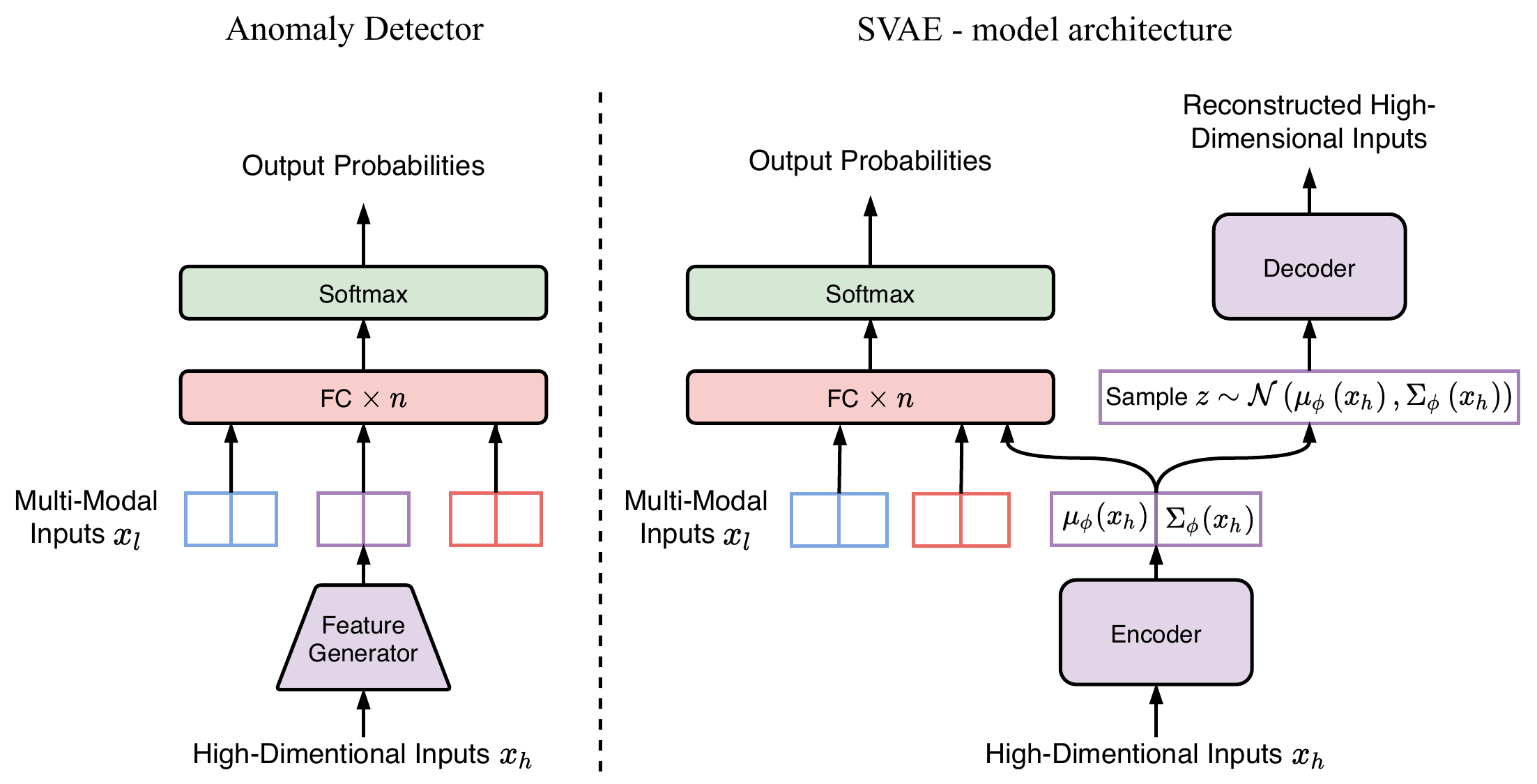}
  \caption{\textit{Left:} The high-dimentional inputs are projected onto a latent space to extract features. The classifier makes an inference based on the compressed representation of the high-dimensional data and other low-dimentional data. \textit{Right:} The VAE is combined with a classifier during training time. The reparameterization trick is omitted.}
  \label{fig:architecture}
\end{figure}

The classification performance of the AD module presented in Figure~\ref{fig:architecture} (left) heavily depends on the quality of the feature generator. The latent variables extracted should be distinct for different classes while preserving the nature of the inputs $\mathbf{x}_h$. We propose a supervised variational autoencoder (SVAE) to unify the training of the FG and the classifier (Figure~\ref{fig:architecture} (right)). During inference, the encoder is used as the FG while the decoder is abandoned.

SVAEs are inspired by the variational autoencoder framework. The decoder uses a generative model of the form:
\begin{equation}
p_\theta(\mathbf{x}_h|\mathbf{z}) = f(\mathbf{x}_h;\mathbf{z}, \theta)
\end{equation}
where $f(\mathbf{x}_h;\mathbf{z}, \theta)$ is a likelihood function whose probability distribution is formed by a non-linear transformation of the latent variable $\mathbf{z}$ with parameters $\theta$. Here, we choose the non-linear transformation to be a multilayer perceptron (MLP) and the likelihood function $f$ to be a Gaussian distribution:
\begin{equation}
f(\mathbf{x}_h;\mathbf{z}, \theta)
=
\mathcal{N} \left( \mathbf{x}_h | \, \text{MLP}(\mathbf{z};\theta),\sigma^2 \cdot I\right)
\end{equation}
where $\text{MLP}(\mathbf{z};\theta)$ is a mean vector, and $\sigma$ is a hyperparameter.

We represent our inference model as an approximate posterior distribution $q_\phi(\mathbf{z}, y|\mathbf{x})$ with variational parameters $\phi$. For the latent variable $\mathbf{z}$, we use a Gaussian inference network as the encoder:
\begin{equation}
q_\phi(\mathbf{z}|\mathbf{x}) =
q_\phi(\mathbf{z}|\mathbf{x}_h) = \mathcal{N}\left(\mathbf{z}|\bm{\mu}_\phi(\mathbf{x}_h), \text{diag}\left(\bm{\Sigma}_\phi(\mathbf{x}_h)\right)\right)
\end{equation}
where $\bm{\mu}_\phi(\mathbf{x}_h)$ is a mean vector, $\bm{\Sigma}_\phi(\mathbf{x}_h)$ is a variance vector, and the functions $\bm{\mu}_\phi: \mathbb{R}^H \mapsto \mathbb{R}^d$ and $\bm{\Sigma}_\phi: \mathbb{R}^H \mapsto \mathbb{R}^d$ are parameterized as MLPs as in the decoder.

The classifier is connected to the output layer of the encoder. We specify this inference model for $y$ as a categorical distribution:
\begin{equation}
q_\phi(y|\mathbf{x}) = \text{Cat}\left(y|\bm{\pi}_\phi\left(\bm{\mu}_\phi(\mathbf{x}_h), \bm{\Sigma}_\phi (\mathbf{x}_h), \mathbf{x}_l \right)\right)
\end{equation}
where $\text{Cat}(y|\bm{\pi}_\phi)$ is the categorical distribution, $\bm{\pi}_\phi(\cdot)$ is a probability vector and is represented as an MLP.

Unlike the prior work in generative semi-supervised model for the digit classification task~\citep{Durk2014} where the inference networks $q_\phi(\mathbf{z}|\mathbf{x})$ and $q_\phi(y|\mathbf{x})$ are assumed to be independent with each other, we expect that the inference model $q_\phi(y|\mathbf{x})$ depends on $q_\phi(\mathbf{z}|\mathbf{x})$. In the former case, the class specification is separated from the writing style of the digit in order to facilitate the generation for specific digits. However, in our case, the features generated by $q_\phi(\mathbf{z}|\mathbf{x})$ relate to classification criteria which are utilized by the classifier to make an inference.

\subsection{Training}
Our model focuses on two different but related tasks: classification and reconstruction. The inference model $q_\phi(y|\mathbf{x})$ generates a probability distribution over class labels $y$ while the generative model $p_\theta(\mathbf{x}_h|\mathbf{z})$ tries to reconstruct the inputs $\mathbf{x}_h$. For our application, the main goal is to find a high-quality inference model for the class label given the current sensor data; however, we found it very useful to learn a generative model simultaneously. We claim that the reconstruction task serves as a regularization~\citep{le2018supervised,liu2016algorithm,caruana1997multitask}, which forces the encoder to learn global features of the high-dimensional inputs that are critical for both the classifier and the decoder. As a positive side effect, by compressing the input dimensions, the model learns to filter out sensor noise and to extract much more robust high-level ideas from those inputs.

Denoting the labeled dataset by $\mathcal{D}$, we specify the overall loss function for SVAE on the entire dataset as:
\begin{equation}
\label{eq.loss-func}
\mathcal{L}
=
\displaystyle\sum_{(\mathbf{x}, y) \in \mathcal{D}}
\left[
-\mathbb{E}_{q_\phi (\mathbf{z}|\mathbf{x})} \left[\log p_\theta(\mathbf{x}_h|\mathbf{z})\right]
+
D_\text{KL}\left[q_\phi (\mathbf{z}|\mathbf{x}) \| p_\theta(\mathbf{z})\right]
-
\alpha \cdot \log q_\phi (y|\mathbf{x})
\right]
\end{equation}
where $p_\theta(\mathbf{z})$ is the prior distribution of the latent variable $\mathbf{z}$, and the hyperparameter $\alpha$ controls the relative weight between the generative and discriminative learning. As in the typical setting of VAE with continuous latent variables~\citep{doersch2016tutorial}, we choose $p_\theta (\mathbf{z})$ as a standard Gaussian distribution $\mathbf{z} \sim \mathcal{N}(0,I)$.

The overall training objective has two parts, each of which focuses on different tasks. The first two terms in equation~(\ref{eq.loss-func}), which is the negative of the evidence lower bound (ELBO) in vanilla VAEs, evaluate the reconstruction loss of the high-dimensional inputs $\mathbf{x}_h$ and contribute to the optimization of the decoder and the encoder. The second KL divergence term can be viewed as a regularization.

The third term in the objective function~(\ref{eq.loss-func}) penalizes the classification error, and contributes to the optimization of the classifier and the encoder. We set $\alpha = 0.1 \cdot N$, where $N$ is the total number of datapoints. An interesting observation from our experiments is that increasing the relative weight $\alpha$ for the classification task does not necessarily generate higher classification accuracy. We argue that with less attention on the reconstruction task, the model tends to lose the generalization property~\citep{le2018supervised}.

The inference model and the generative model can be optimized jointly by stochastic gradient descent of the unified objective function~(\ref{eq.loss-func}). To enable the backpropagation through the sampling process within the network (see Figure~\ref{fig:architecture}), the reparameterization trick is used to move the sampling to a stochastic input layer. We refer readers to~\citep{doersch2016tutorial} for details.

%% file: Sections/04-Experiment.tex
\section{Experimental Results}
\label{sec:experiment}

The proposed model for anomaly detection was evaluated on data collected with the TerraSentia robot in corn and sorghum fields from August to October 2019. The robot navigates through rows of crops under cluttered canopies without damaging the plants. During data collection, the robot was either teleoperated by a human operator or driven by navigation algorithms autonomously while a failure may or may not occur in a run.

In all experiments using SVAEs, we implement the encoder with only one hidden layer and 128 hidden units. The decoder follows the same structure as the encoder. We construct the classifier with one hidden layer and 64 hidden units. We choose a 2-dimensional latent space, $\mathbf{z} \in \mathbb{R}^2$. ReLU activation functions are applied and an Adam optimizer with a constant learning rate of 0.0005 is used to train the network\footnote{The implementation of SVAE is available at \url{https://github.com/tianchenji/Multimodal-SVAE}.}~\citep{kingma2014adam}.

\subsection{TerraSentia Robot and Input Modalities}
TerraSentia is a low-cost, ultracompact and ultralight field robot. The robot is equipped with Hokuyo UST-10LX, a 2D LiDAR sensor which covers a $270^{\circ}$ range with $0.25^{\circ}$ angular resolution, maximum distance reading of 30 m, and 40 Hz update rate. The LiDAR is placed 0.15m to the front of the robot and is adjusted to have a symmetric view about the longitude axis of the robot. The robot is driven by a two-motor set on the left and right wheels. Wheel velocities are measured by a two-channel Hall-effect encoder which provides 64 counts per revolution for each motor.

In our experiments, the AD module takes the LiDAR data of dimension 1080 as the high-dimensional input $\mathbf{x}_h$. The low-dimensional input $\mathbf{x}_l$ has two components:
\begin{inparaenum}[(\itshape i)]
\item
velocities from left and right wheel encoders;
\item
other modalities derived from the raw sensor data.
\end{inparaenum}
In this work, we choose the second part of $\mathbf{x}_l$ as an array consisting of average distances from the robot to the surrounding objects within a set of fixed ranges of scanning angles:
\begin{equation}
\mathbf{x}_l
=
\left[
v_\text{left}; v_\text{right}; \bar{d}_{[\frac{\pi}{3}, \frac{5\pi}{12}]}; \bar{d}_{[\frac{5\pi}{12}, \frac{\pi}{2}]};  \bar{d}_{[\frac{\pi}{2}, \frac{7\pi}{12}]};  \bar{d}_{[\frac{7\pi}{12}, \frac{2\pi}{3}]}
\right]
\in \mathbb{R}^6
\end{equation}
where the subscript of $\bar{d}$ denotes the ranges of LiDAR scanning angle.\footnote{This additional modalities derived from LiDAR might alleviate the reliance on the raw LiDAR data; however, our experiments suggested that the combination of the two generates the best results.} Note that the additional input modalities derived from LiDAR data is the only part that we design specifically for our robotic platform. However, such additional inputs can be designed similarly on other robots without any change of the model architecture.

\subsection{Dataset}
\begin{figure}[t]
  \centering
  \begin{subfigure}[b]{0.27\linewidth}
    \includegraphics[width=\linewidth]{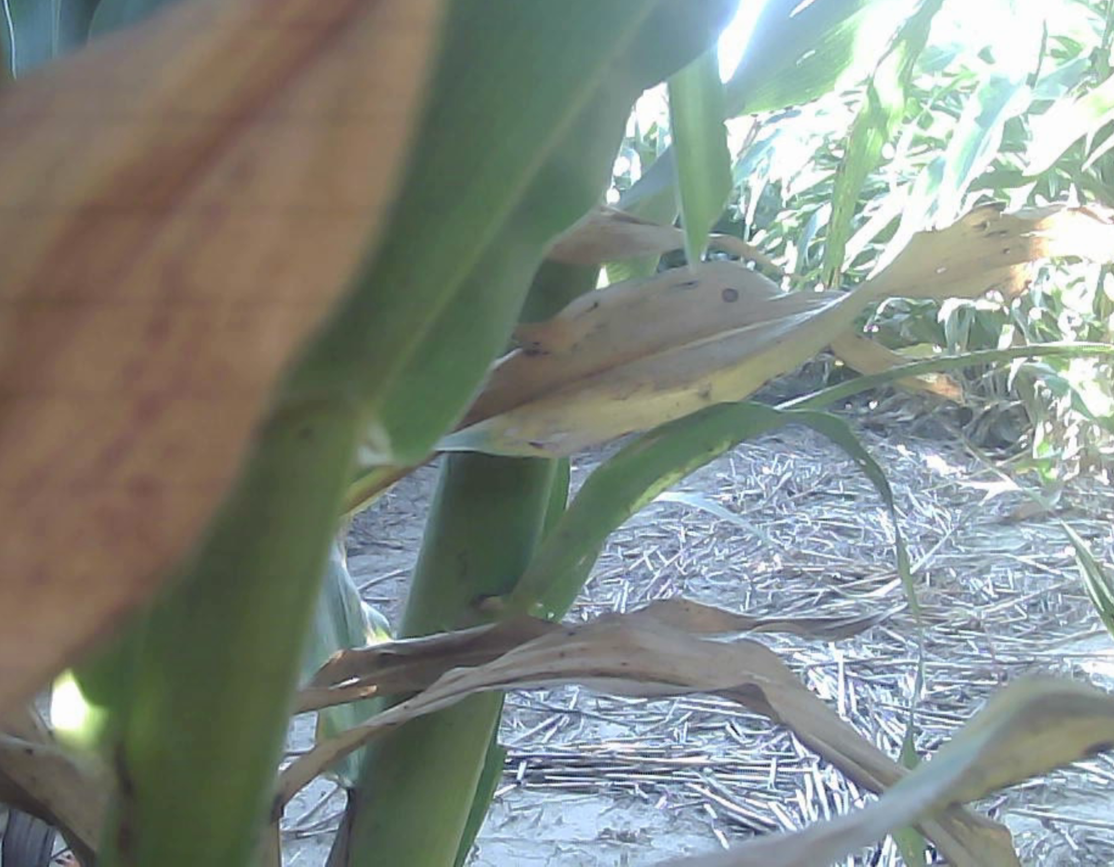}
    \captionsetup{justification=centering}
    \caption{Row collision.}
  \end{subfigure}
  \begin{subfigure}[b]{0.27\linewidth}
    \includegraphics[width=\linewidth]{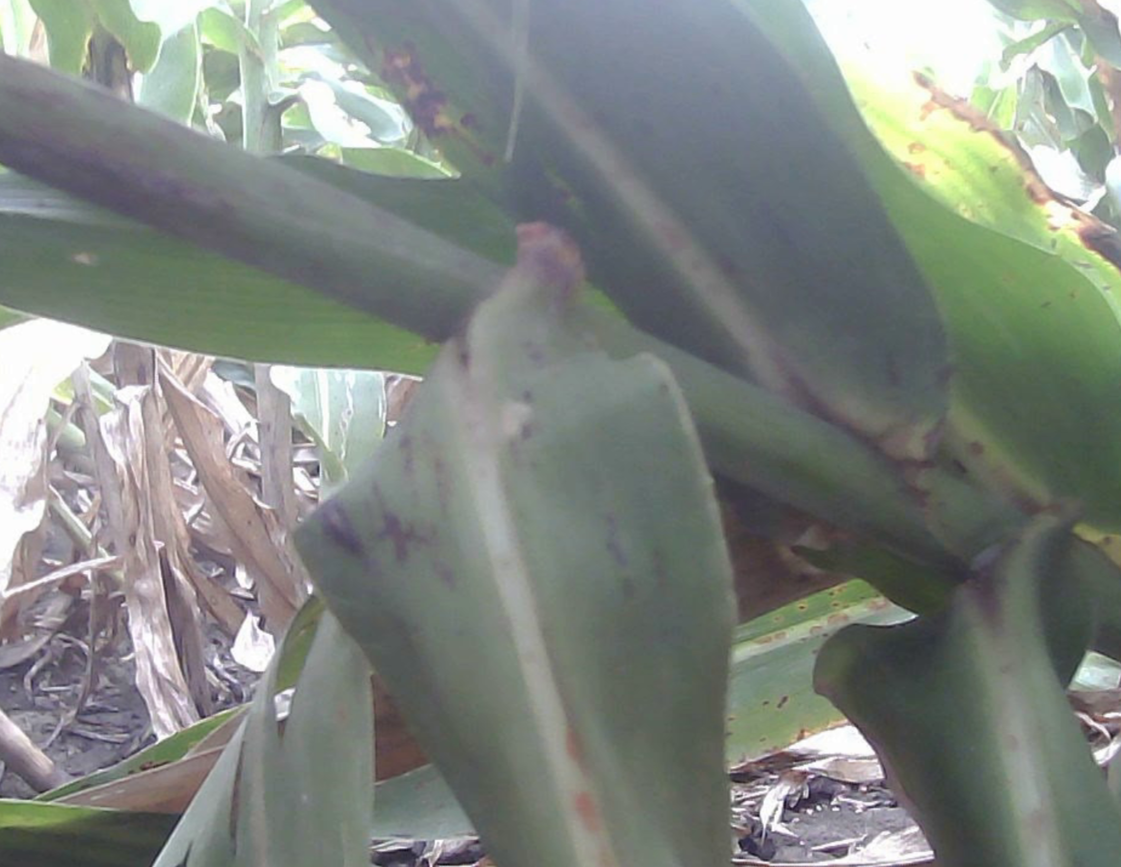}
    \captionsetup{justification=centering}
    \caption{Untraversable obstacle.}
  \end{subfigure}
  \begin{subfigure}[b]{0.27\linewidth}
    \includegraphics[width=\linewidth]{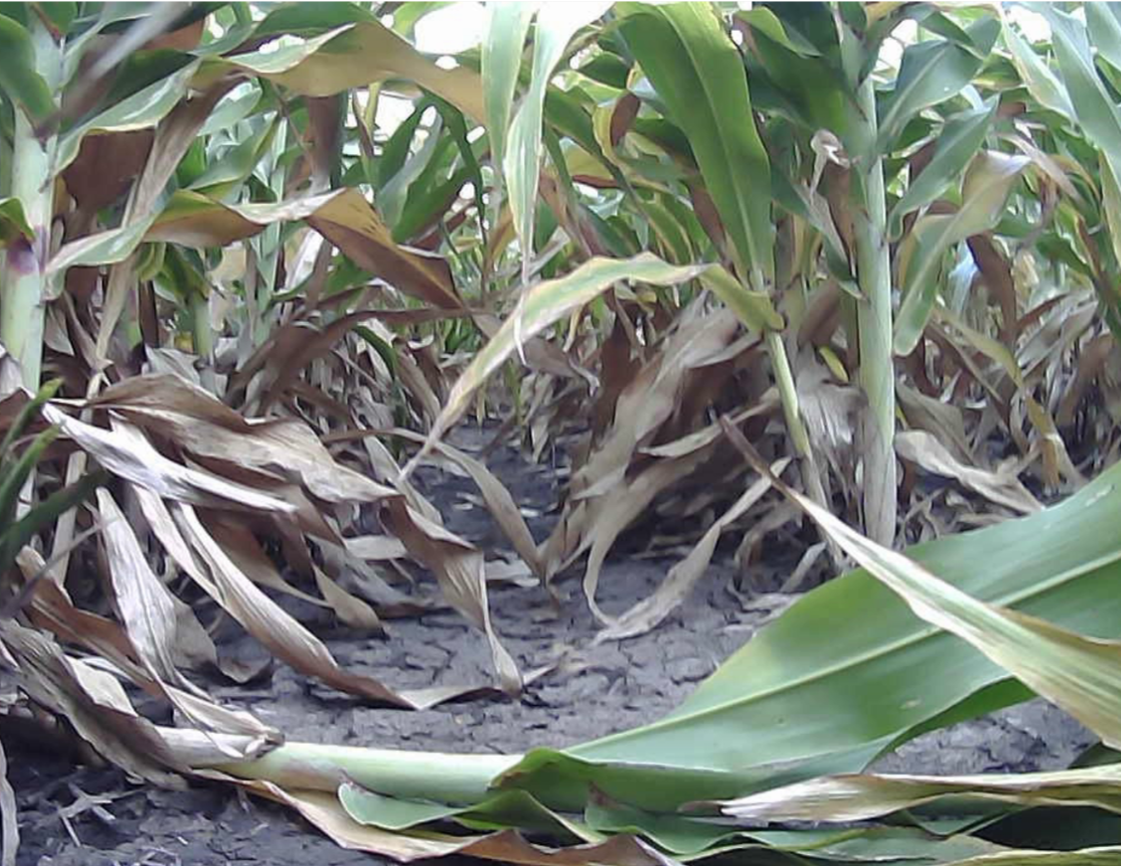}
    \captionsetup{justification=centering}
    \caption{Traversable obstacle.}
  \end{subfigure}
  \captionsetup{justification=centering}
  \caption{Snapshots of anomalous cases from the front camera on the robot in our dataset.}
  \label{fig:data-visuals}
  \vspace{-4mm}
\end{figure}

\vspace{-2mm}
The raw LiDAR data is clipped at 1.8 meters and normalized to [0,1]. We attach a class label to each synchronized data point $\mathbf{x}$. Focusing on the anomalous cases induced by environmental interference, we divide the operation mode during the runs into the following four situations (Figure~\ref{fig:data-visuals}):
\vspace{-1mm}
\begin{enumerate}[label=\arabic*)]
\setcounter{enumi}{-1}
\item
\textit{normal}: The robot follows the center line and navigates safely towards the destination.
\item
\textit{row collision}: The robot deviates from the center line and crashes into crops on either side of the narrow row (lane width of 0.8m).
\item
\textit{untraversable obstacle}: The robot stops in front of the obstacles which obstruct the center line. Such obstacles may have larger size than the robot or are rooted in the ground, making the robot unable to traverse. Typical examples include weeds, lodged plants, etc.
\item
\textit{traversable obstacle}: The robot stops or significantly slows down in front of the obstacles. However, the robot can drive over the obstacle with larger motor torques. Such obstacles may include lodged crop stems, stones, uneven terrains, etc.
\end{enumerate}
\vspace{-1mm}
The above four classes are labeled as an integer from 0 to 3, respectively. Note that each class corresponds to a different recovery maneuver. For instance, the robot can reverse and re-plan the path for row collisions, call for help for untraversable obstacles, or increase motor torque to run over the traversable obstacles. As a result, the output of the AD module can be used as a high-level decision for the planning and control algorithms.

We split the datapoints into a training set (80\%) and a test set (20\%). The training set and test set came from independent runs. Under-sampling of normal cases and over-sampling of anomalous cases were performed to obtain a more balanced dataset for training purposes.
\vspace{-1mm}

\subsection{Results}
\vspace{-2mm}
\paragraph{Failure Identification Performance.}
\label{subsec:results}
We test the performance of SVAE on the test set, along with the following baseline methods:
\vspace{-1mm}
\begin{enumerate}[label=\arabic*)]
\item
MLP: A multilayer perceptron with three hidden layers mapping from multi-modal inputs $\mathbf{x}$ directly to class labels $y$.
\item
PCA + MLR: The high-dimensional input $\mathbf{x}_h$ is compressed through principal component analysis (PCA), then concatenated with low-dimensional input $\mathbf{x}_l$ as the inputs to multinomial logistic regression (MLR).
\item
VAE Fixed Features + MLP: This model was proposed by Kingma et al.~\citep{Durk2014}. The training is a two-stage procedure. First, a VAE is trained by maximizing the ELBO. Then the encoder is used as a feature extractor, and a classifier is trained on the features.
\item
Uni-modal SVAE: The multi-modal inputs $\mathrm{x}_h$ and $\mathrm{x}_l$ are treated identically by concatenating and feeding them to the encoder. The classifier makes an inference merely based on the encoder output. To account for the imbalanced input dimensions, we repeat each element in $\mathrm{x}_l$ by $H$ times to match the size of $\mathrm{x}_h$.
\end{enumerate}

\vspace{-1mm}
Quantitatively, we compare different methods using the following two metrics:
\begin{inparaenum}[(\itshape i)]
\item
precision, including the classification accuracy for each class and the average value;
\item
Kappa coefficient~\citep{cohen1960coefficient}, which is a measure of reliability or agreement between the ground truth and the trained model.\footnote{Kappa coefficient can take any value between -1 to 1. A score of 0 means there is random agreement between the two, whereas a score of 1 means there is a complete agreement among the two. Kappa coefficient is usually regarded as a classification accuracy normalized by the imbalance of the classes in the dataset.}
\end{inparaenum}
\begin{table}[t]
  \vspace{-8mm}
  \begin{center}
  \captionsetup{justification=centering}
    \caption{Classification results with different anomaly detection methods}
    \label{table.comparison}
    \resizebox{\textwidth}{!}{%
    \begin{tabular}{ l | c  c  c  c | c  c }
      \hline
      Model & normal & row collision & \makecell{untraversable \\ obstacle} & \makecell{traversable \\ obstacle} & average & \makecell{Kappa \\ coefficient} \\
      \hline
      \rule{-2.5pt}{2ex} MLP & $98.96\pm0.56$ & $59.04\pm10.50$ & $4.22\pm1.88$ & $69.39\pm13.04$ & $57.90\pm4.47$ & $0.63\pm0.05$ \\
      PCA+MLR & $97.87\pm0.09$ & $78.08\pm0.00$ & $2.22\pm0.00$ & $\mathbf{96.84\pm0.48}$ & $68.75\pm0.11$ & $0.73\pm0.00$ \\
      VAE+MLP & $98.47\pm0.69$ & $79.86\pm6.39$ & $49.89\pm5.98$ & $87.46\pm3.77$ & $78.92\pm1.69$ & $0.81\pm0.03$ \\
      SVAE (uni) & $\mathbf{99.47\pm0.29}$ & $80.41\pm7.91$ & $45.00\pm3.68$ & $84.91\pm1.22$ & $77.45\pm1.87$ & $0.82\pm0.01$ \\
      SVAE & $98.94\pm0.59$ & $\mathbf{82.19\pm6.02}$ & $\mathbf{58.45\pm4.20}$ & $88.42\pm3.50$ & $\mathbf{82.00\pm1.52}$ & $\mathbf{0.84\pm0.02}$ \\
      \hline
    \end{tabular}}
  \end{center}
  \vspace{-3mm}
\end{table}

\begin{figure}[t]
  \centering
  \begin{subfigure}[b]{0.32\linewidth}
    \includegraphics[width=\linewidth]{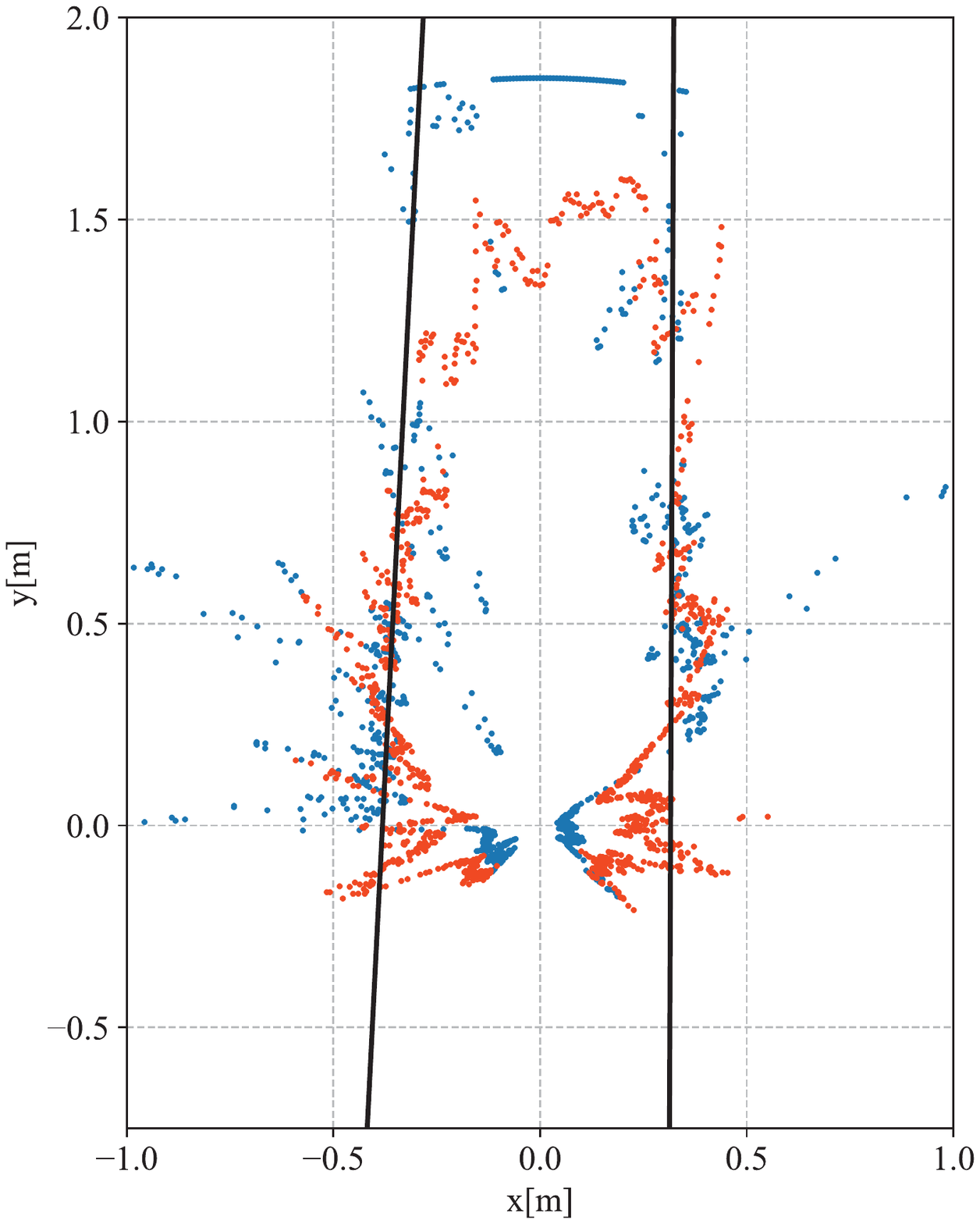}
    \captionsetup{justification=centering}
    \caption{Normal.}
  \end{subfigure}
  \begin{subfigure}[b]{0.32\linewidth}
    \includegraphics[width=\linewidth]{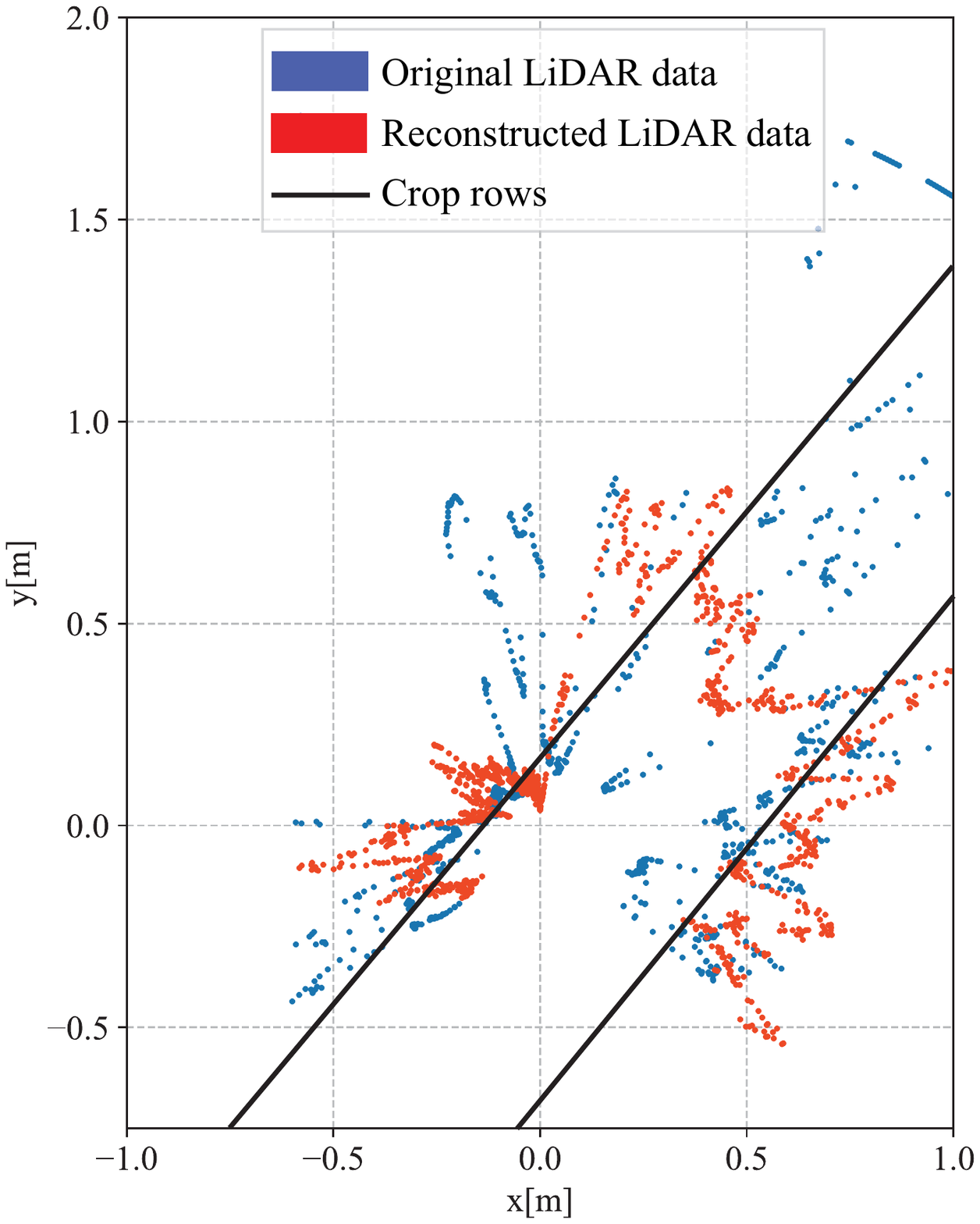}
    \captionsetup{justification=centering}
    \caption{Row collision.}
  \end{subfigure}
  \begin{subfigure}[b]{0.32\linewidth}
    \includegraphics[width=\linewidth]{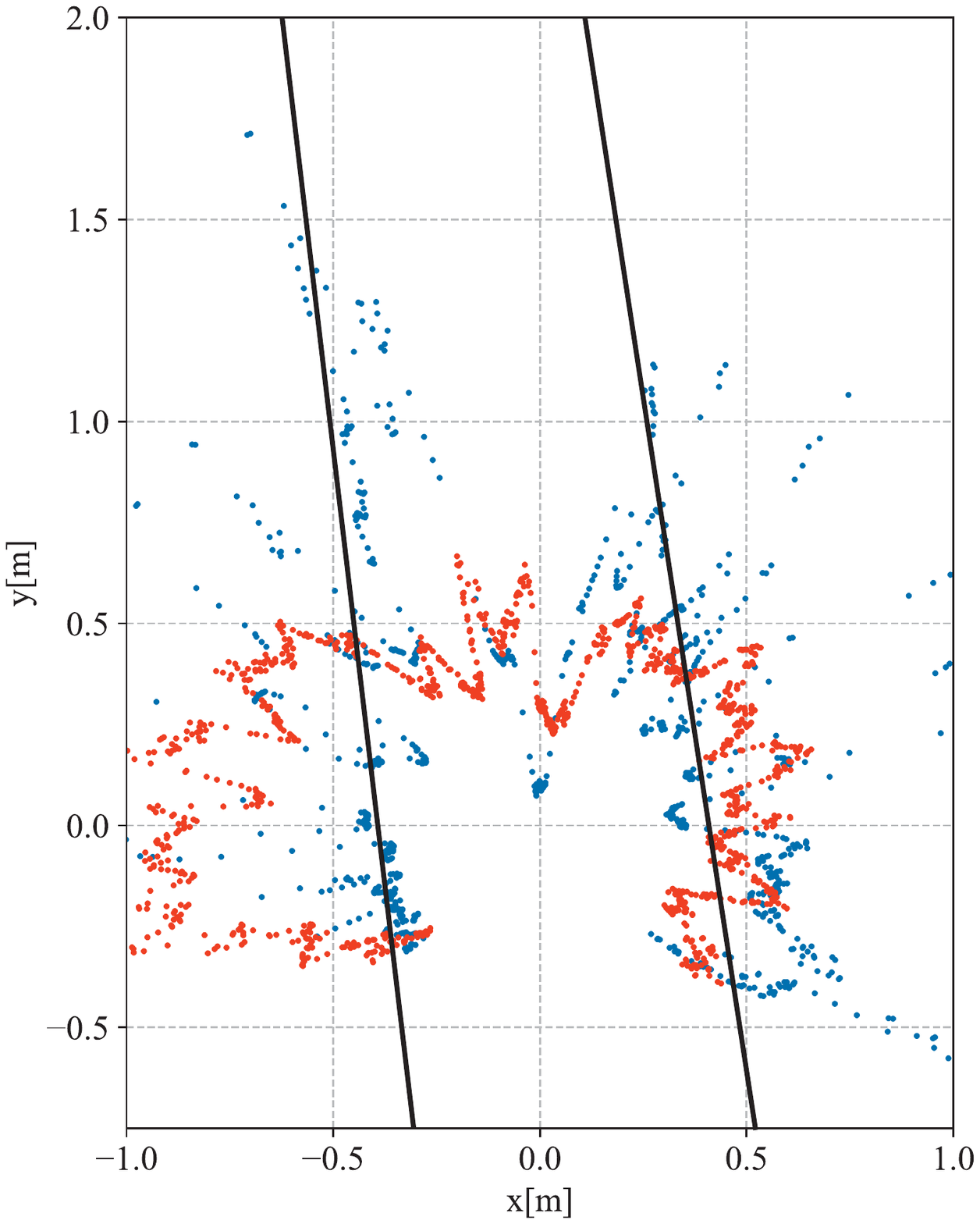}
    \captionsetup{justification=centering}
    \caption{Untraversable obstacles.}
  \end{subfigure}
  \caption{Reconstructed LiDAR point clouds from the latent space. The encoder compresses $\mathbf{x}_h \in \mathbb{R}^{1080}$ to $\mathbf{z} \in \mathbb{R}^2$, then the decoder takes $\mathbf{z}$ to reconstruct $\mathbf{x}_\text{recon} \in \mathbb{R}^{1080}$. The robot is at the origin.}
  \label{fig:recon_lidar_eg}
  \vspace{-4mm}
\end{figure}

All four baselines, along with the SVAE, are evaluated with randomly initialized weights over 10 runs. Table~\ref{table.comparison} summarizes the results. As shown, the SVAE achieves the highest average classificaiton accuracy and best Kappa coefficient. Moreover, the individual classification accuracy of our model for the four classes is either the highest or comparable to the highest among the five models. Note that the MLP generates the overall largest standard deviation whereas PCA+MLR has the lowest standard deviation due to fewer parameters. An interesting observation is that PCA+MLR performs well in classifying three out of the four cases while generating low accuracy for untraversable obstacles. We analysed the confusion matrix and found that $81.11\%$ of the untraversable obstacles were classified as row collision.\footnote{Confusion matrices for all five models are provided in Appendix.} We hypothesize that this is due to ignoring the global features of LiDAR clouds and thus making inference mainly based on local details. The robot should see a tilted path from the LiDAR in the case of row collision, whereas the path should be more aligned with the the robot's longitude axis in the case of untraversable obstacles. As will be shown in Figure~\ref{fig:recon_lidar_eg}, the SVAE learns such global features and thus producing more promising results.

In the SVAE, the encoder is trained to learn the features beneficial to classification and reconstruction, whereas that in VAE+MLP learns the features only useful to the reconstruction task. Such joint learning generally guides SVAE to perform better on classification task. The effectiveness of the multi-modal aspect of our model was verified by the superior classification performance of the SVAE over the uni-modal SVAE. By feeding the inputs $\mathrm{x}_h$ and $\mathrm{x}_l$ to the model at different stages, the encoder is able to focus on learning the features of high-dimensional inputs, without being distracted by low-dimensional input modalities. The relatively large variance of the shape and size of the untraversable obstacles, along with a smallest number of samples in the training set, might be the reasons for its lowest classification accuracy among the four cases.

\paragraph{Learned Feature Space.}
In addition to the quantitative evaluation of our proposed model above, we now present a qualitative results of the learned feature space in the SVAE. The main goal is to visualize what features the latent variable $\mathbf{z}$ is learning from the high-dimensional input $\mathbf{x}_h$.

After training, we feed the data $\mathbf{x}$ in the test set through the model to generate reconstructed LiDAR point clouds using the decoder. Note that the decoder here serves the purpose of qualitative analysis, and is usually ignored in the deployment of the model. Figure~\ref{fig:recon_lidar_eg} shows the results for some typical scenarios. With only two numbers representing the original point cloud of dimension 1080, the decoder can still manage to reconstruct reasonable results, with an average reconstruction error of 0.388m per LiDAR point over the test set. With a more detailed observation, we find that the reconstructions for normal cases and untraversable obstacles are more symmetric about the $y$-axis, while the row collision case shows a more tilted path. Moreover, the reconstructions for the normal cases show more free space in front, whereas that for the untraversable obstacle show less.

\begin{figure}[t]
  \centering
  \begin{subfigure}[b]{0.67\linewidth}
    \includegraphics[width=\linewidth]{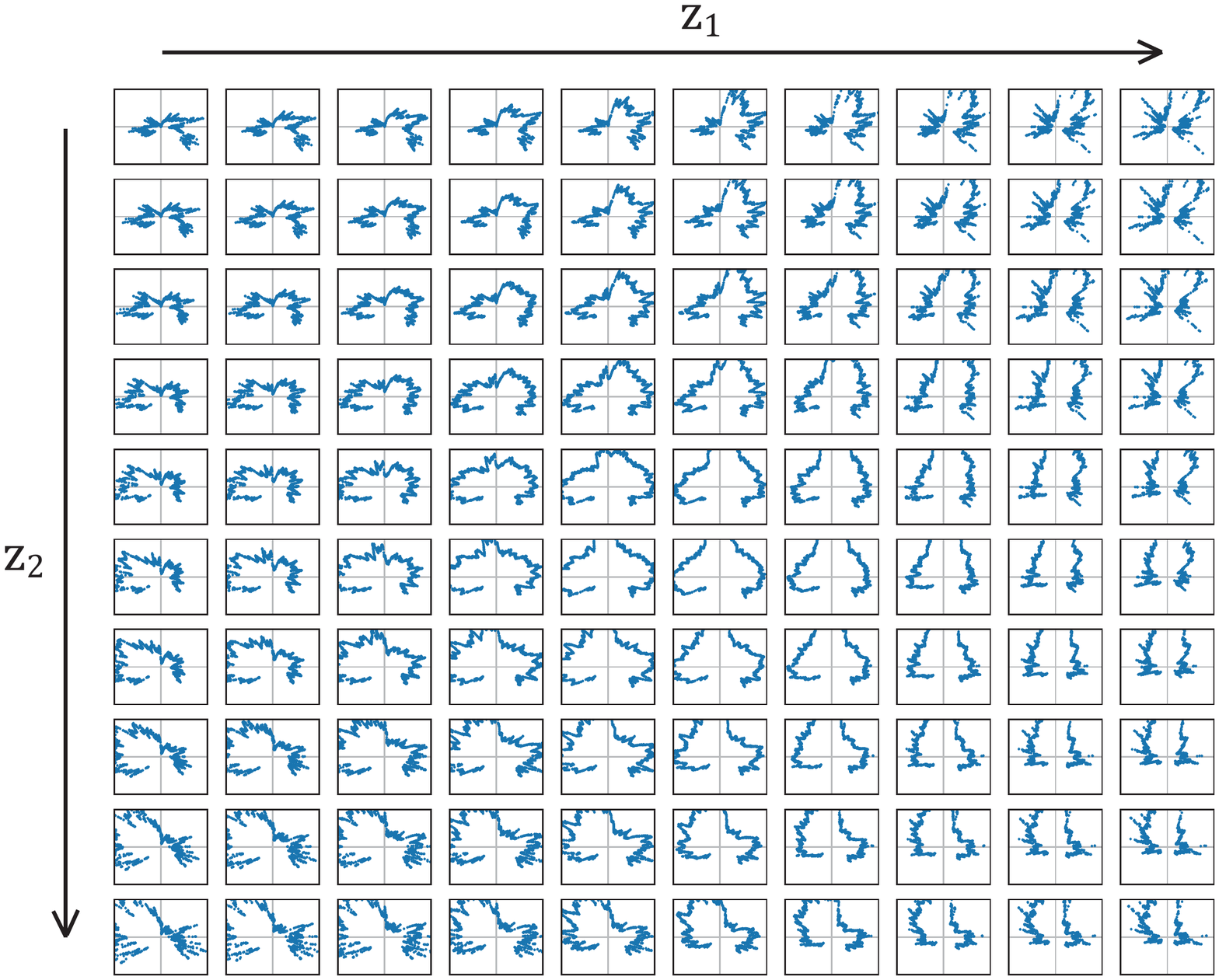}
    \captionsetup{justification=centering}
    \caption{Generated point cloud grid map.}
  \end{subfigure}
  \hspace{0.6cm}
  \begin{subfigure}[b]{0.23\linewidth}
    \includegraphics[width=\linewidth]{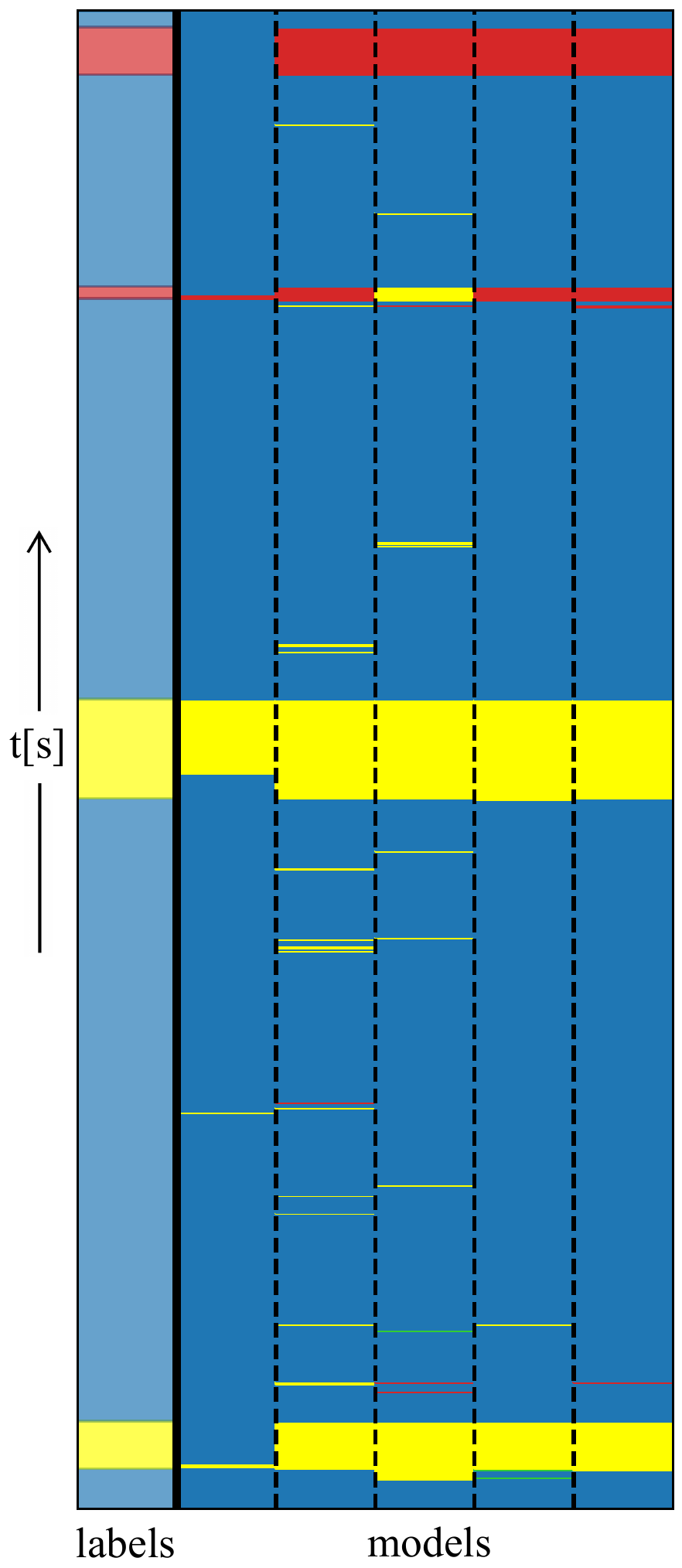}
    \captionsetup{justification=centering}
    \caption{Sensitivity analysis.}
  \end{subfigure}
  \caption{\textbf{(a)} High-level features for LiDAR point clouds obtained by varying latent variables $\mathbf{z}$. \textbf{(b)} The outputs of the five models and ground truth during a run in the test set. The colors of each column represents the actual or inferred labels. We provide more details in Appendix. From left to right: ground truth, MLP, PCA+MLR, VAE+MLP, SVAE (uni-modal), and SVAE. Blue - normal case, Yellow - row collision, Green - untraversable obstacle, red - traversable obstacle.}
  \label{fig:grid-map-sensitivity}
  \vspace{-3mm}
\end{figure}
We further explore the physical meaning of the latent variable by feeding different combinations of $(\mathbf{z}_1, \mathbf{z}_2)$ in the 2-D space to the decoder. Figure~\ref{fig:grid-map-sensitivity} (a) shows the point clouds grid map with $\mathbf{z}_1, \mathbf{z}_2$ increasing in the direction of the arrows. As can be seen, $\mathbf{z}_1$ learns to represent how wide the robot's front view is, making the point clouds in front further away as it increases. In contrast, $\mathbf{z}_2$ manages to represent the orientation of the crop rows, making the point clouds orient from right to left as it increases. The model learns these two important features automatically during training without any specific design or hand tuning.

\paragraph{Sensitivity Analysis.}
In practice, a crucial property of the AD module is its sensitivity to the anomalies. Ideally, the frequency of the declaration of abnormal cases should be similar to that of the actual anomalies. A highly sensitive AD module may frequently intervene during the normal operation of the robot, making it unable to work under uncertain environments with noisy measurements. On the contrary, insensitive AD module may miss some actual failures which could do harm to the robot. In Figure~\ref{fig:grid-map-sensitivity} (b), we investigate the sensitivity by analysing the output of the AD modules in a run. As shown, the SVAE interrupts the robot only when necessary and captures most of the anomalies correctly.

%% file: Sections/05-Conclusion.tex
\section{Conclusion}
\vspace{-2mm}
\label{sec:conclusion}
In this work, we introduced the SVAE, a new family of multi-class classifier for multi-modal failure or anomaly detection which combines VAEs with discriminative models. The training of our model is a simple, one-stage procedure. During test time, high-dimensional modalities are compressed through the encoder to the latent space, and then concatenated with other multi-modal modalities as inputs to the classifier to make an inference. The experiments on field robot data showed that SVAEs are capable of extracting robust and interpretable features from high-dimensional modalities, and that the resultant classifier outperforms baseline methods in failure identification tasks.

%% file: Sections/06-Appendices.tex
\clearpage
\renewcommand\appendixpagename{}

\begin{appendices}
\section{Additional Failure Identification Results}
To further analyze the classification performance of the SVAE along with other baseline methods, we computed the confusion matrices for the five models over the test set. As in Section~\ref{subsec:results}, we compute the mean and standard deviation of the classification results over 10 runs with different weight initializations. Table~\cref{table.confusion-matrix-MLP,table.confusion-matrix-PCA-MLR,table.confusion-matrix-VAE-MLP,table.confusion-matrix-SVAE-uni,table.confusion-matrix-SVAE} summarize the results in percentages.

As shown in the first row in Table~\ref{table.confusion-matrix-MLP}, around $25\%$ of each anomalous case were mis-classified as normal. Without a feature generator extracting features from high-dimensional inputs, MLP struggles with distinguishing generic anomalous cases from the normal case. By contrast, the other four models generate nearly zero probabilities of classifying an abnormal case as normal. This observation indicates the importance of dimensionality compression when dealing with high-dimensional and multi-modal sensor modalities. As mentioned before, PCA+MLR fails to learn the global features of the LiDAR data and thus confuses row collisions and untraversable obstacles. Such claim is confirmed by the similar classification results in the two columns of row collision and untraversable obstacle in Table~\ref{table.confusion-matrix-PCA-MLR}. The confusion matrices for VAE+MLP and SVAE share the most similarities among the five models. However, the SVAE shows the better failure identification performance due to the joint learning of the discriminative model and the generative model. Due to the increased input size and the resultant larger network, the uni-modal SVAE encounters a longer training time and the loss of interpretability of the latent space, without showing a superior classification performance over the SVAE.

\begin{table}[h]
  \begin{center}
  \captionsetup{justification=centering}
    \caption{Confusion matrix for MLP over the test set}
    \label{table.confusion-matrix-MLP}
    \begin{tabular}{ c  l | c  c  c  c }
      \multicolumn{2}{c}{} & \multicolumn{4}{c}{\bfseries Actual} \\
      \multicolumn{2}{c|}{} & normal & row collision & untraversable & traversable \\
      \cline{2-6}
      \rule{-2.5pt}{2ex} \multirow{4}{*}{\bfseries \rotatebox[origin=c]{90}{Predicted}} & normal & $98.96\pm0.56$ & $26.03\pm11.15$ & $24.67\pm6.78$ & $27.54\pm11.53$ \\
      & row collision & $0.72\pm0.39$ & $59.04\pm10.50$ & $69.67\pm6.79$ & $3.07\pm4.01$ \\
      & untraversable & $0.05\pm0.09$ & $0.00\pm0.00$ & $4.22\pm1.88$ & $0.00\pm0.00$ \\
      & traversable & $0.28\pm0.26$ & $14.93\pm6.63$ & $1.44\pm1.57$ & $69.39\pm13.04$ \\
      \cline{2-6}
    \end{tabular}
  \end{center}
  \vspace{-2mm}
\end{table}

\begin{table}[h]
  \vspace{-2mm}
  \begin{center}
  \captionsetup{justification=centering}
    \caption{Confusion matrix for PCA+MLR over the test set}
    \label{table.confusion-matrix-PCA-MLR}
    \begin{tabular}{ c  l | c  c  c  c }
      \multicolumn{2}{c}{} & \multicolumn{4}{c}{\bfseries Actual} \\
      \multicolumn{2}{c|}{} & normal & row collision & untraversable & traversable \\
      \cline{2-6}
      \rule{-2.5pt}{2ex} \multirow{4}{*}{\bfseries \rotatebox[origin=c]{90}{Predicted}} & normal & $97.87\pm0.09$ & $0.00\pm0.00$ & $1.11\pm0.00$ & $0.00\pm0.00$ \\
      & row collision & $1.50\pm0.09$ & $78.08\pm0.00$ & $81.11\pm0.00$ & $3.16\pm0.48$ \\
      & untraversable & $0.25\pm0.00$ & $2.74\pm0.00$ & $2.22\pm0.00$ & $0.00\pm0.00$ \\
      & traversable & $0.37\pm0.00$ & $19.18\pm0.00$ & $15.56\pm0.00$ & $96.84\pm0.48$ \\
      \cline{2-6}
    \end{tabular}
  \end{center}
  \vspace{-2mm}
\end{table}

\begin{table}[h]
  \vspace{-2mm}
  \begin{center}
  \captionsetup{justification=centering}
    \caption{Confusion matrix for VAE+MLP over the test set}
    \label{table.confusion-matrix-VAE-MLP}
    \begin{tabular}{ c  l | c  c  c  c }
      \multicolumn{2}{c}{} & \multicolumn{4}{c}{\bfseries Actual} \\
      \multicolumn{2}{c|}{} & normal & row collision & untraversable & traversable \\
      \cline{2-6}
      \rule{-2.5pt}{2ex} \multirow{4}{*}{\bfseries \rotatebox[origin=c]{90}{Predicted}} & normal & $98.47\pm0.69$ & $0.00\pm0.00$ & $1.11\pm0.00$ & $0.00\pm0.00$ \\
      & row collision & $1.07\pm0.69$ & $79.86\pm6.39$ & $26.67\pm8.36$ & $11.84\pm3.95$ \\
      & untraversable & $0.10\pm0.06$ & $11.78\pm1.16$ & $49.89\pm5.98$ & $0.70\pm1.93$ \\
      & traversable & $0.35\pm0.13$ & $8.36\pm6.07$ & $22.33\pm2.94$ & $87.46\pm3.77$ \\
      \cline{2-6}
    \end{tabular}
  \end{center}
  \vspace{-2mm}
\end{table}

\begin{table}[h!]
  \vspace{-2mm}
  \begin{center}
  \captionsetup{justification=centering}
    \caption{Confusion matrix for SVAE (uni-modal) over the test set}
    \label{table.confusion-matrix-SVAE-uni}
    \begin{tabular}{ c  l | c  c  c  c }
      \multicolumn{2}{c}{} & \multicolumn{4}{c}{\bfseries Actual} \\
      \multicolumn{2}{c|}{} & normal & row collision & untraversable & traversable \\
      \cline{2-6}
      \rule{-2.5pt}{2ex} \multirow{4}{*}{\bfseries \rotatebox[origin=c]{90}{Predicted}} & normal & $99.47\pm0.29$ & $0.00\pm0.00$ & $0.00\pm0.00$ & $0.09\pm0.28$ \\
      & row collision & $0.17\pm0.11$ & $80.41\pm7.91$ & $44.78\pm3.52$ & $14.13\pm0.28$ \\
      & untraversable & $0.18\pm0.21$ & $6.30\pm1.32$ & $45.00\pm3.68$ & $0.88\pm1.09$ \\
      & traversable & $0.18\pm0.07$ & $13.29\pm7.42$ & $10.22\pm0.47$ & $84.91\pm1.22$ \\
      \cline{2-6}
    \end{tabular}
  \end{center}
  \vspace{-3mm}
\end{table}

\begin{table}[ht]
  \vspace{-2mm}
  \begin{center}
  \captionsetup{justification=centering}
    \caption{Confusion matrix for SVAE over the test set}
    \label{table.confusion-matrix-SVAE}
    \begin{tabular}{ c  l | c  c  c  c }
      \multicolumn{2}{c}{} & \multicolumn{4}{c}{\bfseries Actual} \\
      \multicolumn{2}{c|}{} & normal & row collision & untraversable & traversable \\
      \cline{2-6}
      \rule{-2.5pt}{2ex} \multirow{4}{*}{\bfseries \rotatebox[origin=c]{90}{Predicted}} & normal & $98.94\pm0.59$ & $0.00\pm0.00$ & $1.11\pm0.00$ & $0.00\pm0.00$ \\
      & row collision & $0.62\pm0.58$ & $82.19\pm6.02$ & $19.56\pm8.28$ & $11.49\pm3.52$ \\
      & untraversable & $0.08\pm0.03$ & $11.92\pm0.92$ & $58.45\pm4.20$ & $0.09\pm0.28$ \\
      & traversable & $0.36\pm0.15$ & $5.89\pm6.43$ & $20.89\pm5.34$ & $88.42\pm3.50$ \\
      \cline{2-6}
    \end{tabular}
  \end{center}
  \vspace{-4mm}
\end{table}

In practice, robots can always apply larger motor torques to try running over the obstacles without damaging the plants as long as the output of the AD module is not row collision. The robot would either return to normal operation if the obstacle is traversable or call for assistance if the obstacle is untraversable. Such strategy groups traversable and untraversable obstacles as a whole, and thus leads to simplified confusion matrices with higher average classification accuracy of the obstacles than in the original confusion matrices. Here, we only present the results of the SVAE in Table~\ref{table.sim-confusion-matrix-SVAE}, which generates the best classification performance among the five models.
\begin{table}[h]
  \vspace{-3mm}
  \begin{center}
  \captionsetup{justification=centering}
    \caption{Simplified confusion matrix for SVAE over the test set}
    \label{table.sim-confusion-matrix-SVAE}
    \begin{tabular}{ c  l | c  c  c }
      \multicolumn{2}{c}{} & \multicolumn{3}{c}{\bfseries Actual} \\
      \multicolumn{2}{c|}{} & normal & row collision & obstacle \\
      \cline{2-5}
      \rule{-2.5pt}{2ex} & normal & $98.94\pm0.59$ & $0.00\pm0.00$ & $0.49\pm0.00$ \\
      \bfseries Predicted & row collision & $0.62\pm0.58$ & $82.19\pm6.02$ & $15.05\pm3.78$ \\
      & obstacle & $0.44\pm0.16$ & $17.81\pm6.02$ & $84.46\pm3.78$ \\
      \cline{2-5}
    \end{tabular}
  \end{center}
  \vspace{-3mm}
\end{table}
\vspace{-3mm}

\section{Additional Sensitivity Analysis Results}
\begin{figure}[t]
  \centering
  \begin{subfigure}[b]{0.77\linewidth}
    \includegraphics[width=\linewidth]{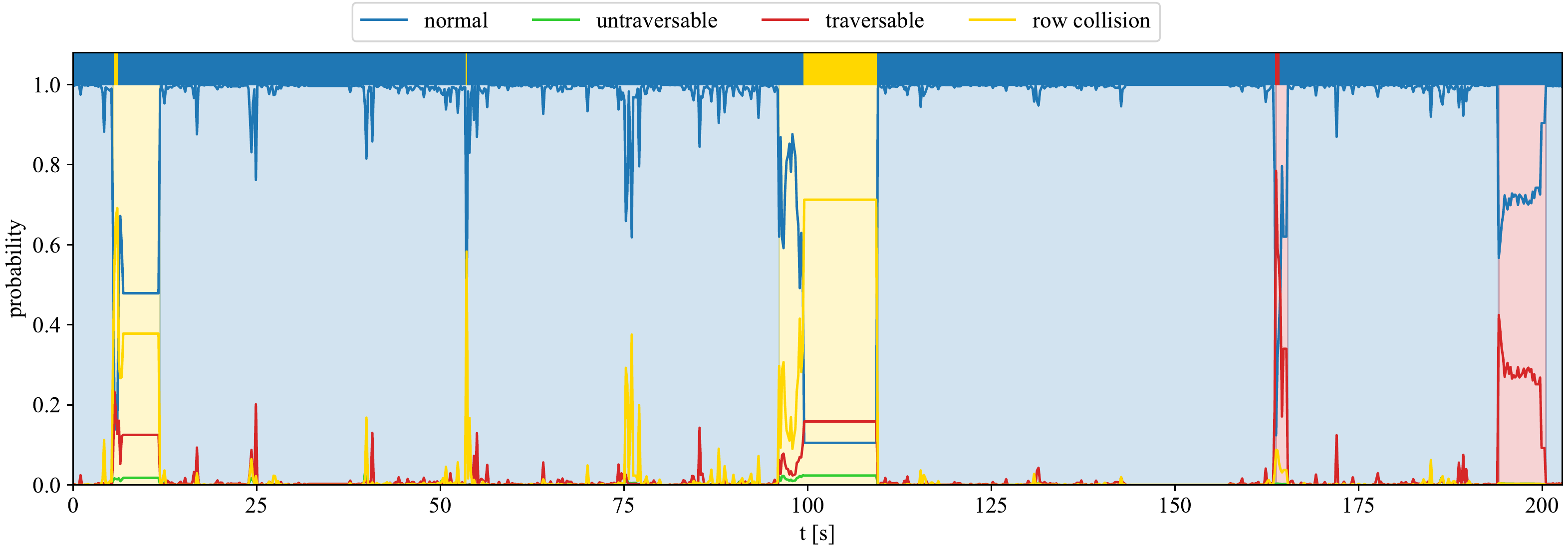}
    \captionsetup{justification=centering}
    \caption{MLP.}
  \end{subfigure}
  \begin{subfigure}[b]{0.77\linewidth}
    \includegraphics[width=\linewidth]{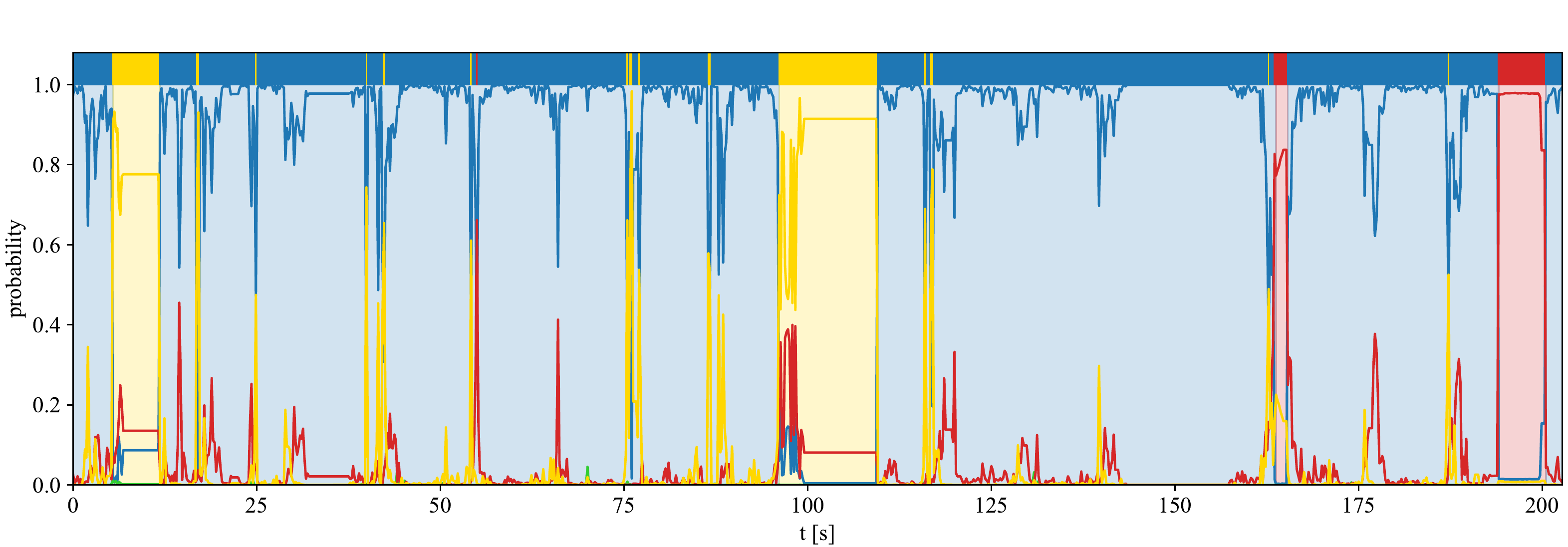}
    \captionsetup{justification=centering}
    \caption{PCA+MLR.}
  \end{subfigure}
  \begin{subfigure}[b]{0.77\linewidth}
    \includegraphics[width=\linewidth]{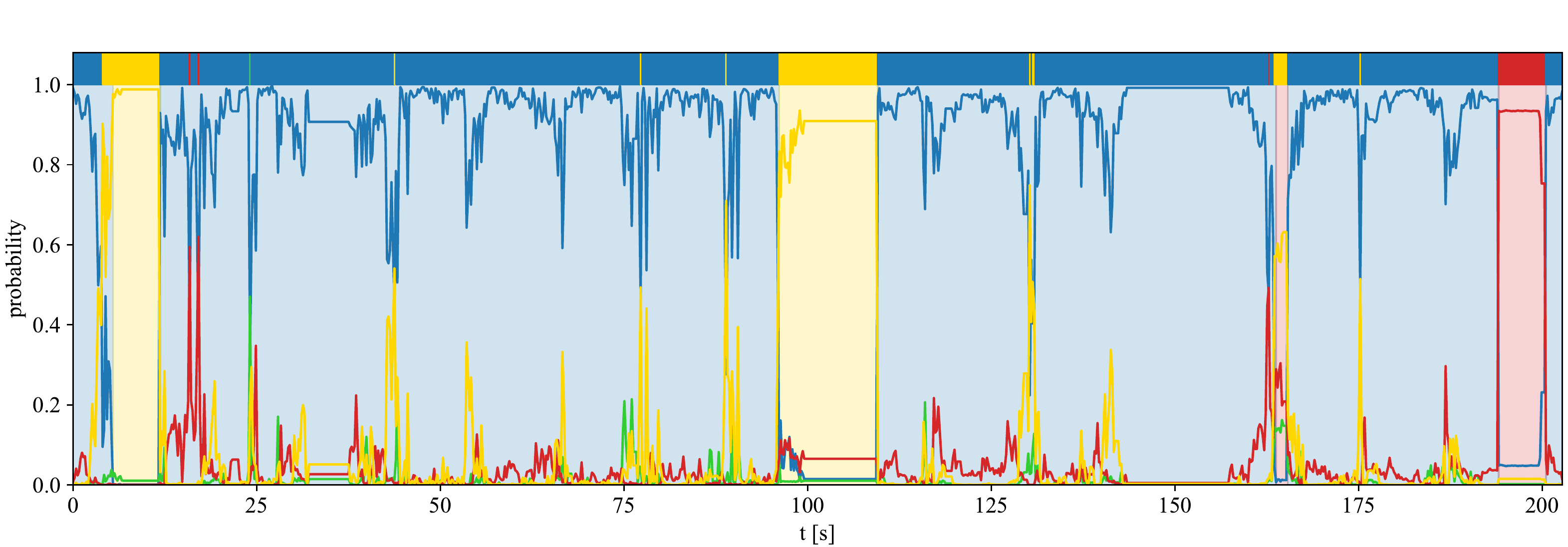}
    \captionsetup{justification=centering}
    \caption{VAE+MLP.}
  \end{subfigure}
  \begin{subfigure}[b]{0.77\linewidth}
    \includegraphics[width=\linewidth]{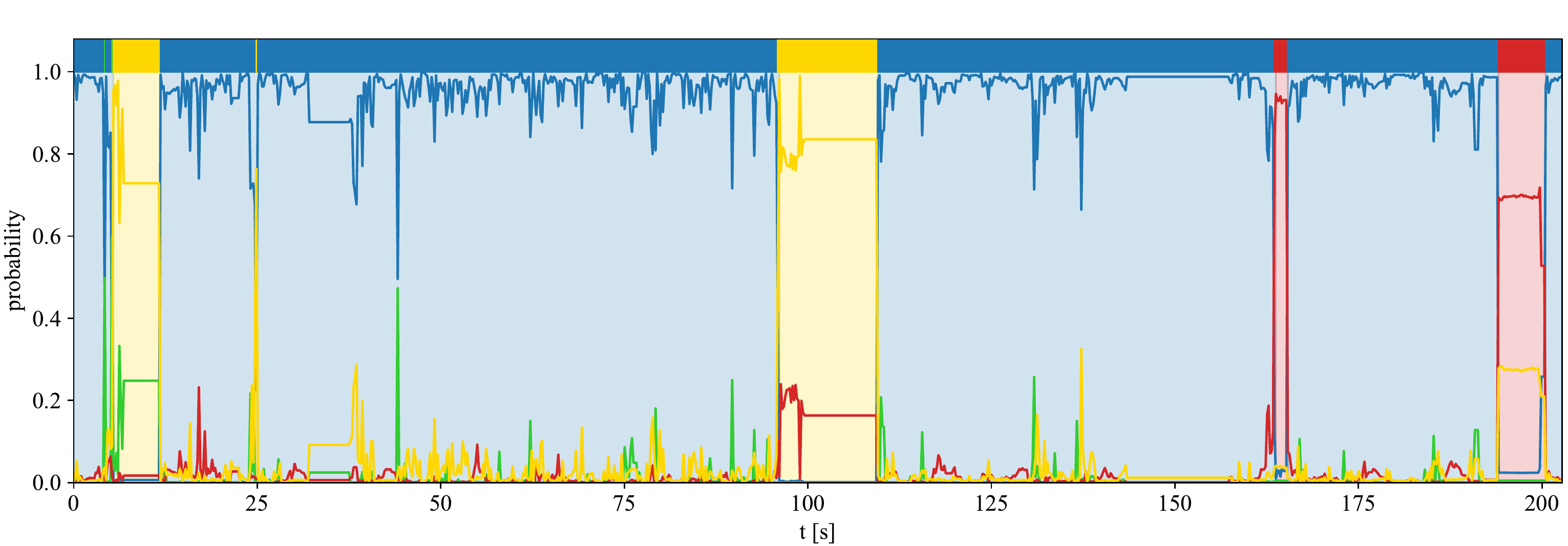}
    \captionsetup{justification=centering}
    \caption{SVAE (uni-modal).}
  \end{subfigure}
  \begin{subfigure}[b]{0.77\linewidth}
    \includegraphics[width=\linewidth]{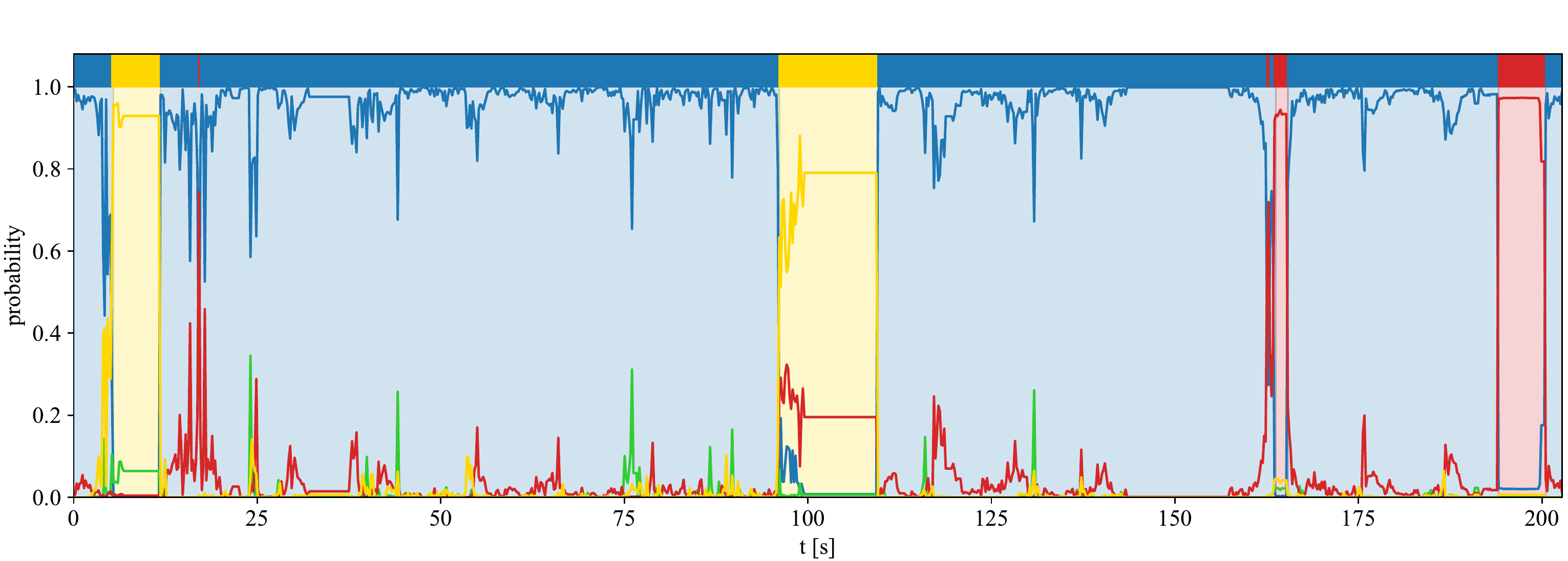}
    \captionsetup{justification=centering}
    \caption{SVAE.}
  \end{subfigure}
  \caption{Detailed sensitivity analysis results. The colors of the top bars represent the outputs (i.e., inferred labels) of different models. The colors of the shaded areas represent the ground truth labels.}
  \label{fig:sensitivity-details}
\end{figure}

Figure~\ref{fig:sensitivity-details} shows the additional results of the sensitivity analysis in Section~\ref{subsec:results} for the five models during the run. The probabilities of classes generated by the models are represented as solid lines. As shown, the uni-modal SVAE and the SVAE generate the best results in this run. With a more detailed observation, the SVAE is more confident in detecting anomalies generally, and has higher classification accuracies for failure cases as suggested by the quantitative results in Section~\ref{subsec:results}.

During data collection, the human operator reset the robot in the correct orientation and resumed moving forward after an anomaly occurred. Such post-anomaly stage before resuming the normal operation is difficult to classify due to the human intervention. To maintain the continuity in time in sensitivity analysis, we keep the output of the AD module unchanged as the previous time step when such discontinuity in time occurs.
\vspace{-1mm}

\section{Baseline Methods Architectures}
We briefly introduce the architectures for the baseline methods implemented in Section~\ref{sec:experiment}:
\vspace{-1mm}
\begin{enumerate}[label=\arabic*)]
\item
MLP: We used three hidden layers. The number of hidden units are $[1024, \, 512, \, 512]$.
\item
PCA+MLR: We compressed the LiDAR data to $30$ components, with $74.48\%$ of variance retained. The classifier is implemented as multinomial logistic regression.\footnote{D. Böhning. Multinomial logistic regression algorithm. \textit{Annals of the Institute of Statistical Mathematics}, 44(1):197-200, 1992.}
\item
VAE+MLP: For the encoder in the VAE, we used the same architecture as in the SVAE with one hidden layer and $128$ hidden units. The decoder mirrors the encoder. We chose a 2-dimensional latent space. The MLP also follows the same architecture as in the SVAE with one hidden layer and $64$ hidden units.
\item
SVAE (uni): To account for the larger input, we used two hidden layers of size $[2048, \, 1024]$ for the encoder. The decoder mirrors the encoder. We chose the latent size to be 128. The classifier has the same structure as in the SVAE with one hidden layer and 64 hidden units.
\end{enumerate}
\vspace{-1mm}
ReLU activation functions were applied. The learning rate for each baseline method was determined individually by selecting the one generating the best average classification accuracy over the four classes, thus may not be strictly the same.

\end{appendices}